\documentclass[journal]{IEEEtran}

\usepackage{cite}
\usepackage{graphicx}
\usepackage{multirow}
\usepackage[pagebackref=true,breaklinks=true,letterpaper=true,colorlinks,bookmarks=false]{hyperref}
\usepackage{amsmath}
\usepackage{amssymb}

\usepackage{pgfplots}
\pgfplotsset{width=9cm, compat=1.9}

\newcommand{\etal}{\textit{et al.}\@ }
\newcommand{\ie}{\textit{i.e.}\@ }

\begin{document}

\title{Simple yet Effective Way for \\ Improving the Performance of GAN
}




\author{Yong-Goo~Shin, Yoon-Jae~Yeo, and Sung-Jea Ko,~\IEEEmembership{Fellow}, IEEE
\thanks{Y.-G. Shin is with School of Electrical Engineering Department, Korea University, Anam-dong, Sungbuk-gu, Seoul, 136-713, Rep. of Korea (e-mail: ygshin@dali.korea.ac.kr).}
\thanks{Y.-J. Yeo is with School of Electrical Engineering Department, Korea University, Anam-dong, Sungbuk-gu, Seoul, 136-713, Rep. of Korea (e-mail: yjyeo@dali.korea.ac.kr).}
\thanks{S.-J. Ko is with School of Electrical Engineering Department, Korea University, Anam-dong, Sungbuk-gu, Seoul, 136-713, Rep. of Korea (e-mail: sjko@korea.ac.kr).}}


\markboth{Submitted to IEEE transactions on Neural Networks and Learning Systems}%
{Shell \MakeLowercase{\textit{Shin et al.}}}

\maketitle

\begin{abstract}
In adversarial learning, discriminator often fails to guide the generator successfully since it distinguishes between real and generated images using silly or non-robust features. To alleviate this problem, this brief presents a simple but effective way that improves the performance of generative adversarial network (GAN) without imposing the training overhead or modifying the network architectures of existing methods. The proposed method employs a novel cascading rejection (CR) module for discriminator, which extracts multiple non-overlapped features in an iterative manner using the vector rejection operation. Since the extracted diverse features prevent the discriminator from concentrating on non-meaningful features, the discriminator can guide the generator effectively to produce the images that are more similar to the real images. In addition, since the proposed CR module requires only a few simple vector operations, it can be readily applied to existing frameworks with marginal training overheads. Quantitative evaluations on various datasets including CIFAR-10, CelebA, CelebA-HQ, LSUN, and tiny-ImageNet confirm that the proposed method significantly improves the performance of GAN and conditional GAN in terms of Frechet inception distance (FID) indicating the diversity and visual appearance of the generated images.


\end{abstract}

\section{Introduction}
\label{sec1}
\IEEEPARstart{G}{enerative} adversarial network (GAN)~\cite{goodfellow2014generative} based on deep convolutional neural networks (CNNs) has shown considerable success to capture complex and high-dimensional image data, and been utilized to numerous applications including image-to-image translation~\cite{isola2017image, choi2018stargan, zhu2017unpaired}, image inpainting~\cite{yu2018free, shin2020pepsi++, sagong2019pepsi}, and text-to-image translation~\cite{reed2016generative, hong2018inferring}. Despite the recent advances, however, the training of GAN is known to be unstable and sensitive to the choices of hyper-parameters~\cite{zhang2018self}. To address this problem, some researchers proposed the novel generator and discriminator structures~\cite{karras2017progressive, zhang2018self, zhang2018stackgan++}. These methods effectively improves image generation task on challenging datasets such as ImageNet~\cite{krizhevsky2012imagenet} but are difficult to apply to other various applications since they impose training overhead or need to modify the network architectures.  

Several works~\cite{arjovsky2017wasserstein, gulrajani2017improved, miyato2018spectral, odena2017conditional, mao2019mode, chen2019self, deshpande2018generative, wu2019sliced, deshpande2019max} attempted to stabilize the training of GAN by using novel loss functions or regularization terms. Arjovsky~\etal~\cite{arjovsky2017wasserstein} applied the Wasserstein distance to adversarial loss function, which shows better training stability than the original loss function. Gulrajani~\etal~\cite{gulrajani2017improved} extended the method in~\cite{arjovsky2017wasserstein} by adding the gradient regularization term, called gradient penalty, to further stabilize the training procedure. Recently, some researchers~\cite{deshpande2018generative, wu2019sliced, deshpande2019max} proposed sliced Wasserstein distance-based adversarial loss function which employs the multiple projection vectors in the last layer to handle diverse feature spaces. However, these methods require many projection vectors to capture important feature; these techniques have to keep unuseful projection vectors since they do not distinguish between meaningful and non-meaning projection vectors. 

Apart from these approaches, Miyato~\etal~\cite{miyato2018spectral} proposed the weight normalization technique, called spectral normalization, which limits the spectral norm of weight matrices to stabilize the training of the discriminator. Furthermore, by combining the projection discriminator~\cite{miyato2018cgans} with the spectral normalization, they significantly improved the performance of the conditional image generation task on ImageNet~\cite{krizhevsky2012imagenet}. Recently, Chen~\etal~\cite{chen2019self} combined GAN with self-supervised learning by adding an auxiliary loss function. Although this method improves the performance of GAN, it needs an additional task-specific network and objective functions for self-supervised learning, which results in extra computational loads on training procedure. Mao~\etal~\cite{mao2019mode} proposed a simple regularization term which maximizes the ratio of the distance between images with respect to the distance between latent vectors. This regularization term does not impose the training overhead and not require the network structure modification, which makes it readily applicable to various applications. 

Inspired by the method in~\cite{mao2019mode}, this brief presents a simple yet effective way that greatly improves the performance of GAN without modifying the original network architectures or imposing the training overhead. In general, the discriminator extracts features using multiple convolutional layers and predicts whether the input image is real or fake using a fully connected layer which produces a single scalar value. Indeed, the operation of fully connected layer computing the single scalar value is equivalent to the inner product operation. In other words, the last layer produces the scalar value by conducting the inner product between a single embedding vector, \ie a weight vector, and an image feature vector obtained via the multiple convolutional layers. In the inner product process, however, the discriminator mainly considers a particular direction of feature vector; the discriminator unintentionally ignores the part of the feature space which is perpendicular to the weight vector. Since the generator is trained through adversarial learning which focuses on deceiving the discriminator, it produces an image without considering the ignored feature space. For instance, if the discriminator first learns the global structure for distinguishing between the real and generated images, the generator will naturally attempt to produce the images having a similar global structure with real images without considering local structure. In other words, the generator fails to fully capture the complex and high-dimensional feature space on the image data. 

To alleviate this problem, we propose a novel cascading rejection (CR) module which extracts different features in an iterative procedure using the vector rejection operation. Since the extracted non-overlapped features prevent the discriminator from concentrating on a single particular feature, the discriminator can guide the generator effectively to produce the images that are more similar to the real images. In addition, since the proposed CR module requires only a few simple vector operations, it can be readily employed to existing frameworks with marginal training overheads. We conducted extensive experiments on various datasets including CIFAR-10~\cite{27torralba200880}, CelebA~\cite{liu2015deep}, Celeb-HQ~\cite{karras2017progressive, liu2015deep}, LSUN~\cite{yu15lsun}, and tiny-ImageNet~\cite{23deng2009imagenet, yao2015tiny}. Experimental results show that the proposed method significantly improves the performance of GAN and conditional GAN in terms of a Frechet inception distance (FID) indicating the diversity and visual appearance of the generated images. 

In summary, in this brief we present:
\begin{itemize}
\item A simple but effective technique for improving the performance of GAN without imposing the training overhead and modifying the network structures. 
\item A novel CR module which prevents the discriminator from focusing on the non-meaningful feature for distinguishing between real and generated images. 

\end{itemize}


\section{Preliminaries}
\label{sec2}
\subsection{Generative adversarial network}
\label{sec:2.1.GAN}

Typically, GAN~\cite{goodfellow2014generative} consists of the generator $G$ and the discriminator $D$. In GAN, both networks are simultaneously trained: $G$ is trained to create a new image which is indistinguishable from real images, whereas $D$ is optimized to differentiate between real and generated images. This relation can be considered as a two-player min-max game where $G$ and $D$ compete with each other. Formally, the $G$~($D$) is trained to minimize~(maximize) the loss function, called adversarial loss, which is expressed as follows:
\begin{eqnarray}
\label{eq1:gan}
    \lefteqn{\min_G \max_D E_{x\sim P_\textrm{data}(x)}[\log D(x)]}\nonumber\\
    & & {} +E_{z\sim P_{z(z)}}[\log(1-D(G(z)))],
\end{eqnarray}\\
where $z$ and $x$ denote a random noise vector and a real image sampled from the noise $P_z(z)$ and real data distribution $P_\textrm{data}(x)$, respectively. It is worth noting that $D(x)$ and $D(G(z))$ are scalar values indicating the probabilities that $x$ and $G(z)$ came from the data distribution. 

Conditional GAN (cGAN) which aims at producing the class conditional images have been actively researched~\cite{mirza2014conditional, odena2017conditional, miyato2018spectral, zhang2017stackgan}. cGAN usually adds conditional information $c$, such as class labels or text condition, to both generator and discriminator in order to control the data generation process in a supervised manner. This can be formally expressed as follows:
\begin{eqnarray}
\label{eq2:cgan}
    \lefteqn{\min_G \max_D E_{(x,c)\sim P_{\mathrm{data}}(x)}[\log D(x,c)]}\nonumber\\
    && {} +E_{z\sim P_{z(z)},c\sim P_{\mathrm{data}}(x)}[\log(1-D(G(z,c)))]
\end{eqnarray}\\
By training the networks based on the above equation, the generator can select an image category to be generated, which is not possible when employing the GAN framework.

\subsection{Revisit the Fully Connected Layer}
\label{sec:2.2.FC}
To train the discriminator using Eqs.~\ref{eq1:gan} and~\ref{eq2:cgan}, the discriminator should produce the single scalar value as an output. To this end, in the last layer, the discriminator usually employs a fully-connected layer with a single output channel; the fully-connected layer acts like the inner product between an embedding vector \textbf{w}, \ie weight vector, and the image feature vector \textbf{v} obtained through the multiple convolutional layers. Even if the last layer consists of several pixels such as PatchGAN~\cite{isola2017image}, the discriminator conducts the inner product for each pixel and averages all values for the adversarial loss function. 

\begin{figure}[t]
\centering
\includegraphics[width=0.8\linewidth]{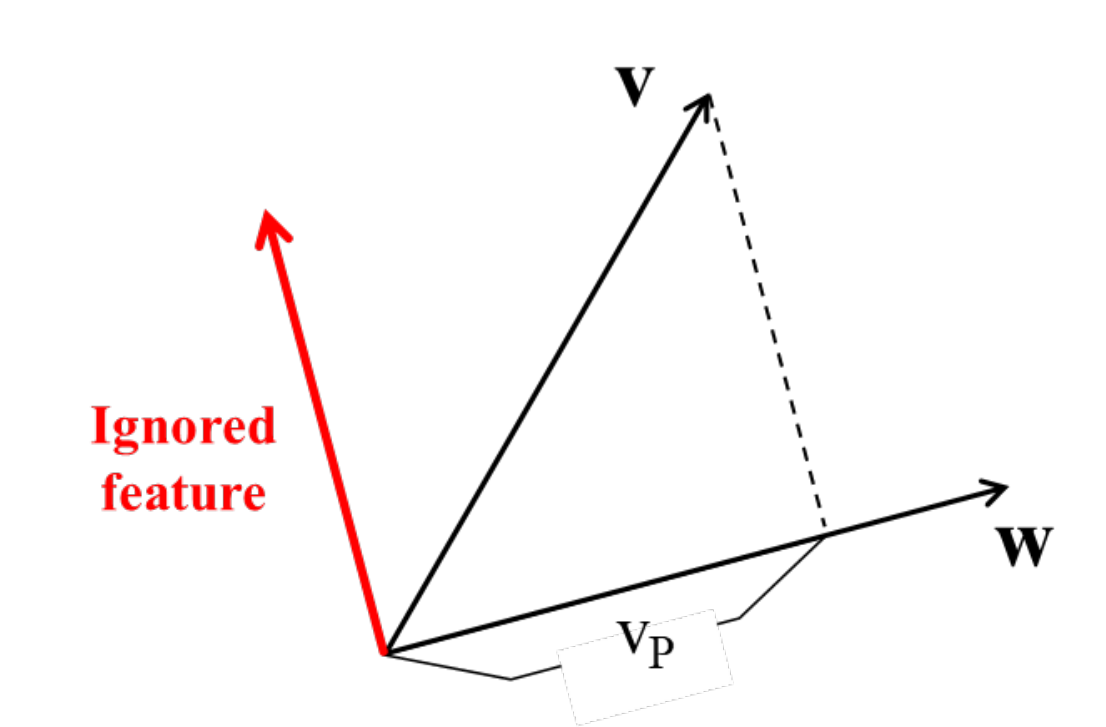}
\caption{Example of the inner product. In the inner product process, the feature space which is perpendicular to \textbf{w} is ignored.}
\label{fig:fig1}
\end{figure}

\begin{figure*}[!ht]
\centering
\includegraphics[width=0.9\linewidth]{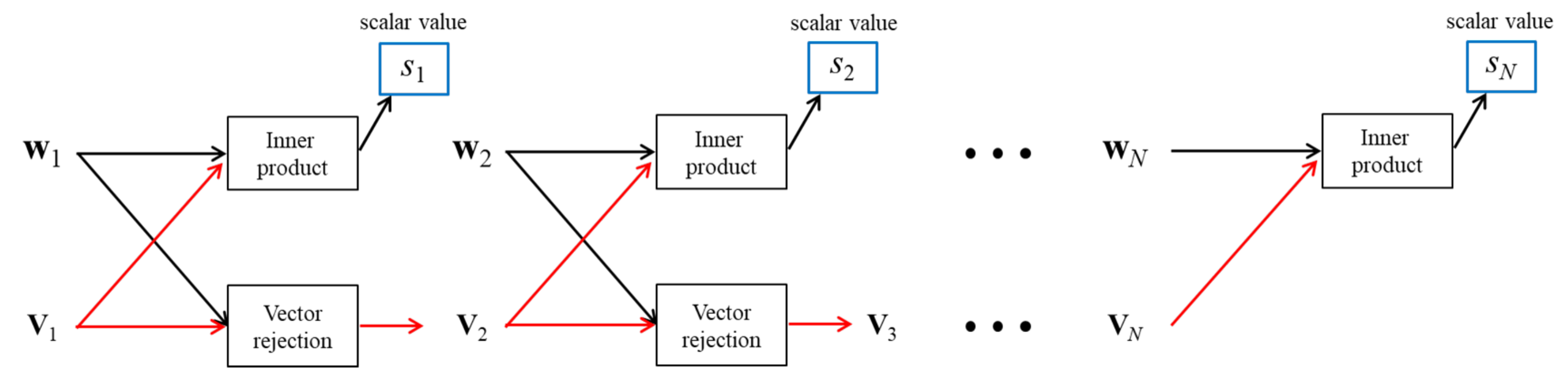}
\caption{The illustration of the CR module. In the CR module, \textit{N} scalar values are obtained through iterative vector rejection and inner product processes.}
\label{fig:fig2}
\vspace{-0cm}
\end{figure*}

The inner product of \textbf{v} onto \textbf{w} is illustrated in Fig.1. As shown in Fig.~\ref{fig:fig1}, the inner product produces the scalar value V$_\textrm{P}$, but it ignores the part of the feature space which is perpendicular to the \textbf{w}. In other words, the discriminator mainly considers the feature space which is parallel to the \textbf{w} when computing the scalar value for the adversarial loss. This problem often makes the discriminator difficult to successfully guide the generator. To alleviate this problem, this brief proposes the CR module which encourages the discriminator to consider ignored feature space in the last layer.


\section{Proposed Method}
\subsection{Cascading rejection module}
\label{subsec:3.1}
In Euclidean space, the fully-connected layer producing a single scalar value as an output is equivalent to the inner product \textit{P}(\textbf{v}, \textbf{w}) as follows: 

\begin{equation}
\label{eq3:IP}
    P(\textrm{\textbf{v}}, \textrm{\textbf{w}}) = \textrm{\textbf{v}}^\textrm{T}\textrm{\textbf{w}}
\end{equation}\\
where \textrm{\textbf{v}} and \textrm{\textbf{w}} indicate the input feature vector and the embedding vector, \ie weight vector, in the last fully connected layer, respectively. From this formulation, we observe that the ignored feature caused by the inner product, \textbf{\^{v}}, can be obtained by the vector rejection of \textrm{\textbf{v}} from \textrm{\textbf{w}}, which is defined as follows:

\begin{equation}
\label{eq4:RJ}
    \textbf{\^{v}} = \textrm{\textbf{v}} - \frac{\textrm{\textbf{w}}^\textrm{T}\textrm{\textbf{v}}}{\textrm{\textbf{w}}^\textrm{T}\textrm{\textbf{w}}}\textrm{\textbf{w}}.
\end{equation}\\
In other words, by minimizing (or maximizing) the adversarial loss using the additional scalar value obtained through the inner product of \textbf{\^{v}} and another weight vector \textbf{\^{w}}, the discriminator is able to consider the ignored feature space. 

Based on these observations, we propose the CR module which iteratively conducts the inner product and vector rejection processes. Fig.~\ref{fig:fig2} illustrates the proposed CR module, where \textbf{v}$_1$ indicates the input feature vector of the CR module, which is obtained through the multiple convolutional layers in the discriminator. The iterative vector rejection process generates $N-1$ vectors, \ie $\textbf{v}_i,\{i=2,\dots, N\}$, which represent the ignored feature in the previous inner product operation. The iterative inner product operation produces $N$ scalar values, \ie $s_i,\{i=1,\dots , N\}$, which is used to distinguish the real and generated samples. Note that $\textbf{v}_i$s have different directions with each other since they are obtained through the vector rejection operation. By using the scalar values obtained via the CR module, the adversarial loss of the discriminator $L_D$ and that of the generator $L_G$ can be rewritten as 

\begin{equation}
\label{eq5:LD}
L_D=-E[\frac{1}{N}\displaystyle\sum_{i=1}^{N}log(\sigma(s_{x,i}))] - E[\frac{1}{N}\displaystyle\sum_{i=1}^{N}(1 - \log(\sigma(s_{z,i})))],
\end{equation}

\begin{equation}
\label{eq6:LG}
    L_G=E[\frac{1}{N}\displaystyle\sum_{i=1}^{N}(1 - \log(\sigma(s_{z,i})))],
\end{equation}\\
where $s_{x,i}$ ($s_{z,i}$) indicates the \textit{i}-th scalar value when the input of discriminator is real (generated) image, and $\sigma(\cdot)$ represents a sigmoid function. 

When \textit{N} is one, the loss functions in Eqs.~\ref{eq5:LD} and~\ref{eq6:LG} are equivalent to the original one in Eq.~\ref{eq1:gan}. In contrast, when \textit{N} is larger than one, the discriminator and generator should consider the ignored feature space to minimize the $L_D$ and $L_G$, respectively. It is worth noting that since the iterative inner product and vector rejection processes are simple vector operations, the proposed CR module does not impose the training overhead. In addition, since the proposed CR module is appended after the last fully connected layer in the discriminator, there is no modification to the existing architecture of the discriminator.

\begin{figure*}[t]
\centering
\includegraphics[width=0.9\linewidth]{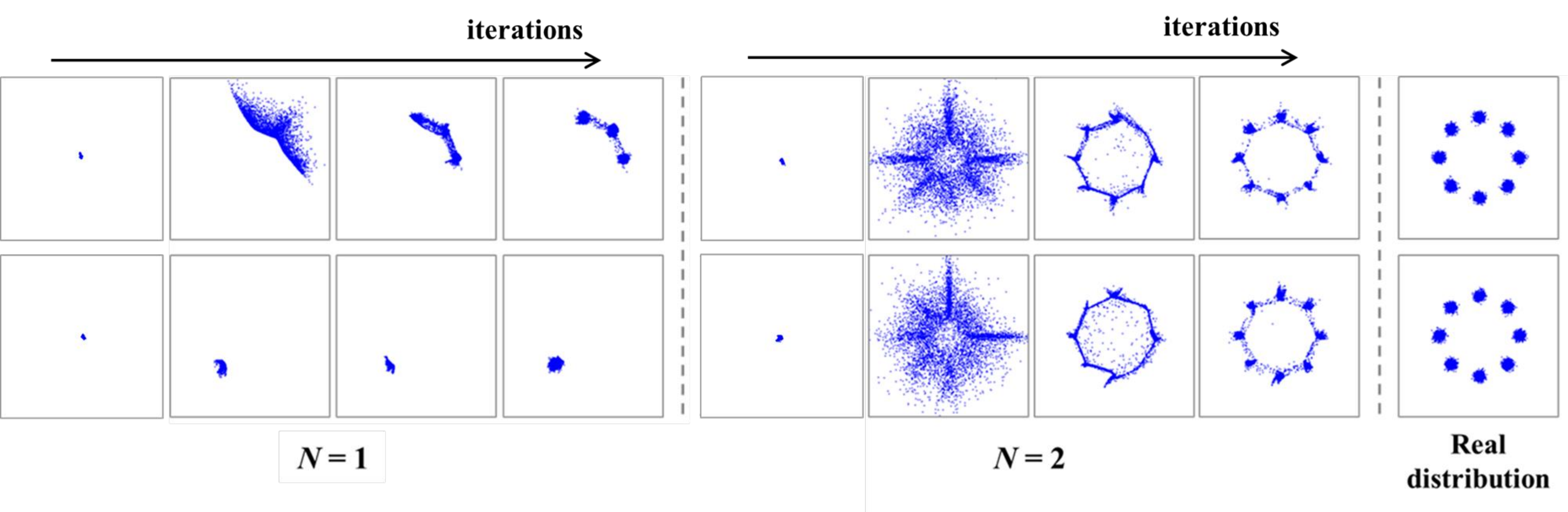}
\caption{The illustration of the experimental results on eight 2D Gaussian mixture models. We trained the networks two times to reveal the true trend of the CR module.}
\label{fig:fig3}
\end{figure*}

\subsection{Understanding and analysis}
\label{subsec:3.2}
In this subsection, we demonstrate the theoretical validity of the proposed method. Let us consider the discriminator $D(x; \theta_D)$ which produces a feature vector $\textbf{v}_1$, \textit{i.e.} the input of the CR module, where $x$ and $\theta_D$ indicate the input and learning parameters of the discriminator, respectively. As mentioned in the previous subsection, the CR module produces \textit{N} output values. For instance, when $N=2$, the output values, $s_1$ and $s_2$, are obtained as follows:

\begin{equation}
\label{eq7:s1}
    s_1 = \textbf{w}_1^\textrm{T}\textbf{v}_1,
\end{equation}

\begin{equation}
\label{eq8:s2}
    s_2 = \textbf{w}_2^\textrm{T}\textbf{v}_2 = \textbf{w}_2^\textrm{T}(\textbf{v}_1 - \frac{\textbf{w}_1^\textrm{T}\textbf{v}_1}{\textbf{w}_1^\textrm{T}\textbf{w}_1}\textbf{w}_1).
\end{equation}

By using $s_1$ and $s_2$, the adversarial loss $L_{adv}$ can be defined as $f(s_1) + f(s_2)$, where $f(\cdot)$ is a function corresponding to the divergence of distance measure of the user's choice. In the training process, the derivative of $L_{adv}$ with respect to $\theta_D$ is computed as follows:

\begin{equation} 
\label{eq9}
\begin{split}
\frac{\partial L_{adv}}{\partial \theta_D} & = (\frac{\partial f(s_1)}{\partial \textbf{v}_1} + \frac{\partial f(s_2)}{\partial \textbf{v}_1})\frac{\partial \textbf{v}_1}{\partial \theta_D} \\
 & = [f'(s_1)\textbf{w}_1 + f'(s_2)(\textbf{w}_2 - \frac{\textbf{w}_1^\textrm{T}\textbf{w}_2}{\textbf{w}_1^\textrm{T}\textbf{w}_1}\textbf{w}_1)]\frac{\partial \textbf{v}_1}{\partial \theta_D}.
\end{split}
\end{equation}\\
In Equation~\ref{eq9}, the first term $f'(s_1)\textbf{w}_1$ is equivalent to the derivative of the traditional adversarial loss, whereas the second term represents the derivative resulted from the CR module. It is worth noting that the second term is identical to the vector rejection of $\textbf{w}_2$ from $\textbf{w}_1$; the second term has orthogonal direction with the first term. Since the second term guides the discriminator to consider a new direction which is perpendicular to $\textbf{w}_1$, it prevents the discriminator from focusing on the one particular direction, \ie $\textbf{w}_1$, during the training procedure. In other words, without any constraints on $\textbf{w}_1$ and $\textbf{w}_2$, the second term prevents the discriminator from from concentrating on a single feature which cannot guide the generator well. To show the ability of the proposed method, we have trained GAN with/without the CR module on eight 2D Gaussian mixture models (GMMs) using simple network architectures consisting of multiple fully-connected layers. Figure~\ref{fig:fig3} shows the experimental results. As training progresses, the traditional GAN ($N=1$) suffers from the mode collapse problem, whereas the proposed method learns all GMMs successfully. 

\begin{figure}[t]
\centering
\includegraphics[width=0.85\linewidth]{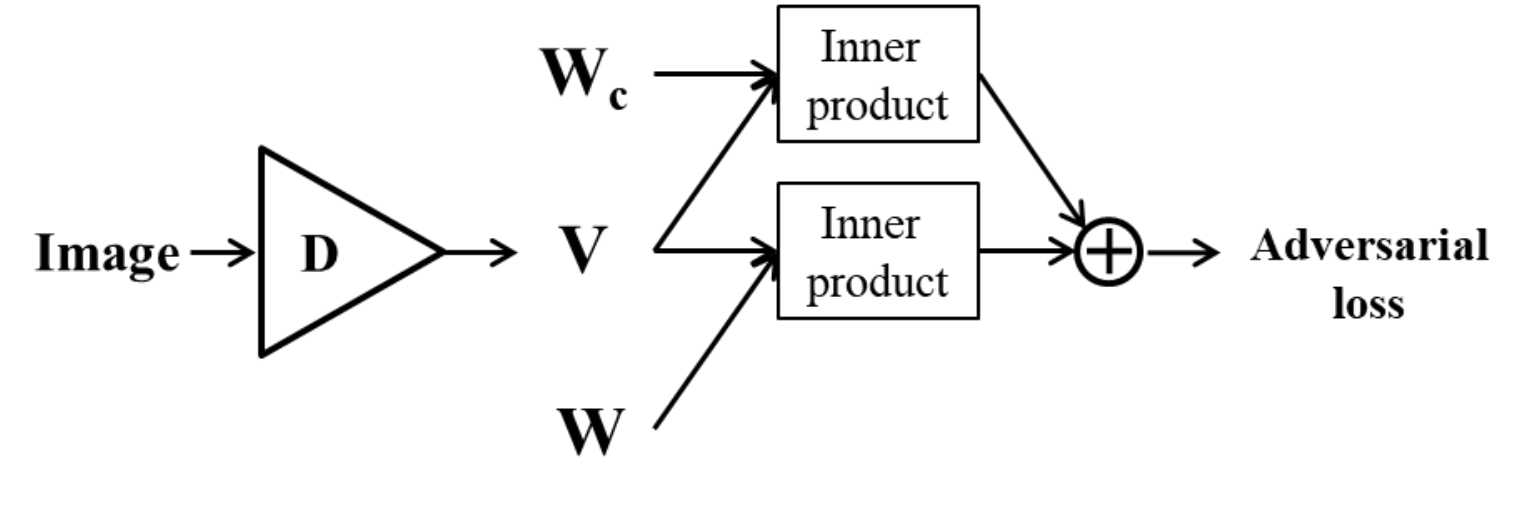}
\caption{The illustration of the conditional projection  discriminator~\cite{miyato2018cgans}.}
\label{fig:fig4}
\end{figure}

\subsection{Conditional Cascading Rejection module}
\label{subsec:3.3}
To apply the CR module for cGAN scheme, in this subsection, we propose the CR module for the cGAN, called the conditional cascading rejection (cCR) module. Among the various cGAN frameworks~\cite{mirza2014conditional, reed2016generative, salimans2016improved, odena2016semi, odena2017conditional, miyato2018cgans}, we design the cCR module based on the conditional projection discriminator~\cite{miyato2018cgans} which shows superior performance than other existing discriminators for the cGAN. As depicted in Fig.~\ref{fig:fig4}, the conditional projection discriminator takes an inner product between the embedded condition vector \textbf{w}$_\textrm{c}$, which is a different vector depending on the given condition, and the feature vector of the discriminator, \ie\textbf{v}, so as to impose a regularity condition. Based on the regularity condition, the conditional projection discriminator produces the conditional scalar value \textit{S}$_\textrm{c}$(\textbf{v}, \textbf{w}, \textbf{w}$_\textrm{c}$) as follows:

\begin{equation}
\label{eq8:Pc}
    S_c(\textrm{\textbf{v}}, \textrm{\textbf{w}}, \textrm{\textbf{w}}_c) = \textrm{\textbf{v}}^\textrm{T}\textrm{\textbf{w}} + \textrm{\textbf{v}}^\textrm{T}\textrm{\textbf{w}}_c = \textrm{\textbf{v}}^\textrm{T}(\textrm{\textbf{w}}+\textrm{\textbf{w}}_c).
\end{equation}

\begin{figure}[t]
\centering
\includegraphics[width=1\linewidth]{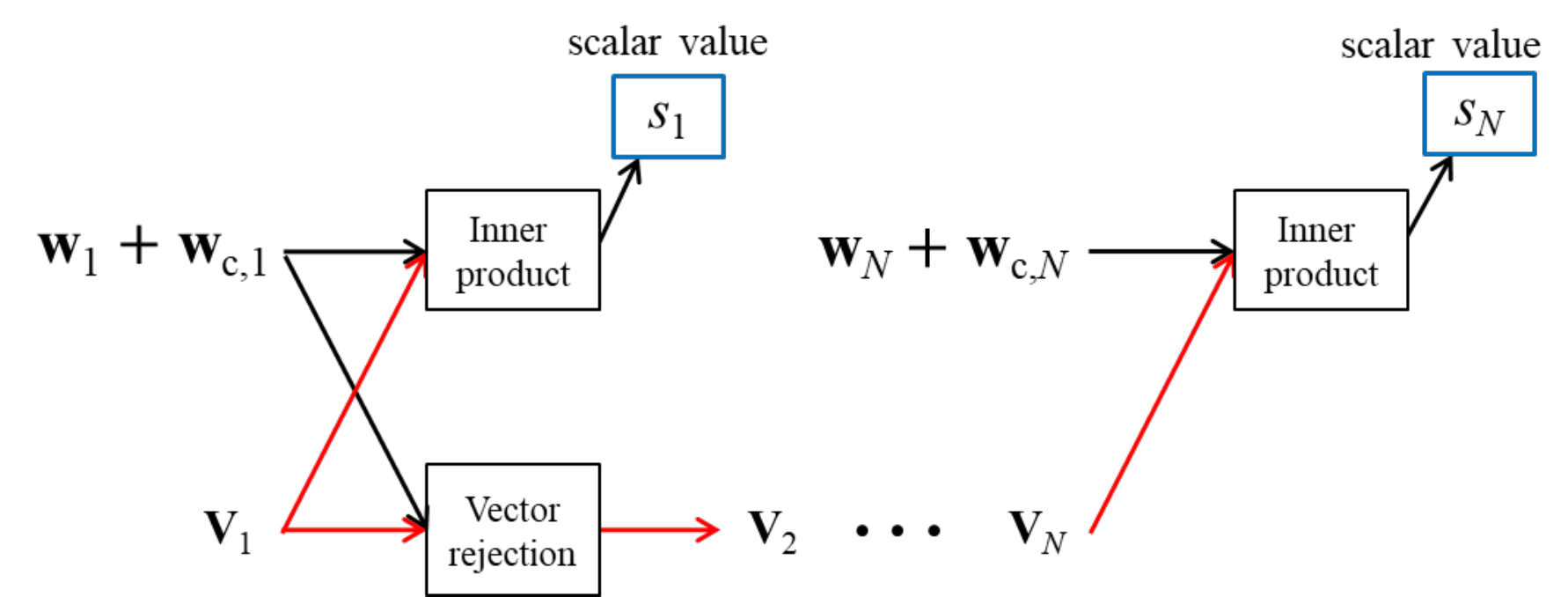}
\caption{The illustration of the cCR module. In this brief, we propose the cCR based on the conditional projection discriminator in~\cite{miyato2018cgans}.}
\label{fig:fig5}
\vspace{+0.1cm}
\end{figure}

Based on the above equation, as shown in Fig.~\ref{fig:fig5}, we design the cCR module by replacing the \textbf{w}$_i$ in the CR module with (\textbf{w}$_i$ + \textbf{w}$_{\textrm{c}, i}$). More specifically, the $i$-th vector rejection process of the cCR module can be expressed as follows:

\begin{equation}
\label{eq8_2:Rc}
    \textbf{v}_{i+1} = \textrm{\textbf{v}}_i - \frac{(\textrm{\textbf{w}}_i+\textbf{w}_{c, i})^\textrm{T}\textrm{\textbf{v}}_i} {(\textrm{\textbf{w}}_i+\textbf{w}_{c, i})^\textrm{T}(\textrm{\textbf{w}}_i+\textbf{w}_{c, i})}(\textrm{\textbf{w}}_i+\textbf{w}_{c, i}).
\end{equation}\\
Unlike the CR module for GAN, the cCR module conducts the vector rejection by considering the conditional information. After producing \textit{N} scalar values through the cCR module, the conditional adversarial loss can be easily computed by using Eqs.~\ref{eq5:LD} and~\ref{eq6:LG}.

\begin{table*}[t]
\caption{Detailed architectures of generator according to the image resolution.}
\begin{center}
\begin{tabular}{c | c | c | c | c}
\hline\hline
$32\times32$ resolution & $64\times64$ resolution & $128\times128$ resolution & $256\times256$ resolution & $512\times512$ resolution \\
\hline
$z \in \mathbb{R}^{128} \sim N(0, I)$ & $z \in \mathbb{R}^{128} \sim N(0, I)$ & $z \in \mathbb{R}^{128} \sim N(0, I)$ & $z \in \mathbb{R}^{256} \sim N(0, I)$ & $z \in \mathbb{R}^{256} \sim N(0, I)$\\
FC, $4 \times 4 \times 256$ & FC, $4 \times 4 \times 1024$ & FC, $4 \times 4 \times 1024$ & FC, $4 \times 4 \times 512$ & FC, $4 \times 4 \times 512$ \\
ResBlock, up, 256 & ResBlock, up, 512 & ResBlock, up, 512 & ResBlock, up, 512 & ResBlock, up, 512\\
ResBlock, up, 256 & ResBlock, up, 256 & ResBlock, up, 256 & ResBlock, up, 512 & ResBlock, up, 512\\
ResBlock, up, 256 & ResBlock, up, 128 & ResBlock, up, 128 & ResBlock, up, 256 & ResBlock, up, 256\\
BN, ReLU & ResBlock, up, 64 & ResBlock, up, 64 & ResBlock, up, 128 & ResBlock, up, 128\\
$3\times3$ conv, Tanh & BN, ReLU & ResBlock, up, 32 & ResBlock, up, 64 & ResBlock, up, 64\\
 & $3\times3$ conv, Tanh & BN, ReLU & ResBlock, up, 32 & ResBlock, up, 32\\
& & $3\times3$ conv, Tanh & BN, ReLU  & ResBlock, up, 16\\
& & & $3\times3$ conv, Tanh  & BN, ReLU \\
& & & & $3\times3$ conv, Tanh   \\

\hline\hline
\end{tabular}
\end{center}
\label{table:table1}
\end{table*}

\begin{table*}[t]
\caption{Detailed architectures of discriminator according to the image resolution.}
\begin{center}
\begin{tabular}{c | c | c | c | c}
\hline\hline
$32\times32$ resolution & $64\times64$ resolution & $128\times128$ resolution & $256\times256$ resolution & $512\times512$ resolution \\
\hline
RGB image & RGB image & RGB image & RGB image & RGB image \\
ResBlock, down, 128 & ResBlock, down, 64 & ResBlock, down, 32 & ResBlock, down, 32 & ResBlock, down, 16\\
ResBlock, down, 128 & ResBlock, down, 128 & ResBlock, down, 64 & ResBlock, down, 64 & ResBlock, down, 32\\
ResBlock, 128 & ResBlock, down, 256 & ResBlock, down, 128 & ResBlock, down, 128 & ResBlock, down, 64\\
ResBlock, 128 & ResBlock, down, 512 & ResBlock, down, 256 & ResBlock, down, 256 & ResBlock, down, 128\\
ReLU & ResBlock, 512 & ResBlock, down, 512 & ResBlock, down, 512 & ResBlock, down, 256\\
Global sum pooling & ReLU & ResBlock, 512 & ResBlock, down, 512 & ResBlock, down, 512\\
CR module, \textit{N} & Global sum pooling & ReLU & ResBlock, 512 & ResBlock, down, 512\\
 & CR module, \textit{N} & Global sum pooling & ReLU & ResBlock, 512\\
 & & CR module, \textit{N} & Global sum pooling & ReLU\\
 & &  & CR module, \textit{N} & Global sum pooling\\
 & &  & & CR module, \textit{N} \\

\hline\hline
\end{tabular}
\end{center}
\label{table:table2}
\end{table*}


\subsection{Implementation details}
\label{sec4.1}
In order to evaluate the effectiveness of the CR module, we conducted extensive experiments using the CIFAR-10~\cite{27torralba200880}, LSUN~\cite{yu15lsun}, CelebA~\cite{liu2015deep}, and tiny-ImageNet~\cite{23deng2009imagenet, yao2015tiny} datasets. The CIFAR-10 and LSUN datasets consist of 10 classes, whereas the tiny-ImageNet, which is a subset of the ImageNet~\cite{23deng2009imagenet}, is composed of 200 classes. Among a large number of images in the LSUN dataset, we randomly selected 30,000 images for each class; we employed 300,000 images for training. In addition, we resized the images from the CelebA and LSUN datasets as $64\times 64$ pixels, whereas the tiny-ImageNet dataset as $128\times 128$ pixels. To train the network producing the high-resolution image, we employed the CelebA-HQ dataset~\cite{liu2015deep, karras2017progressive} by resizing the image into $256\times256$ and $512\times512$ resolutions. For the objective function, we adopted the \textit{hinge} version of adversarial loss. The \textit{hinge} version loss with the CR module can be expressed as

\begin{equation}
    L_D=E[\frac{1}{N}\displaystyle\sum_{i=1}^{N}\max(0, 1-s_{x,i})] + E[\frac{1}{N}\displaystyle\sum_{i=1}^{N}max(0, 1+s_{z,i})],
\label{eq.eq10}
\end{equation}

\begin{equation}
    L_G=-E[\frac{1}{N}\displaystyle\sum_{i=1}^{N}s_{z,i}].
\label{eq.eq11}
\end{equation}

\begin{figure}[t]
\centering
\includegraphics[width=0.85\linewidth]{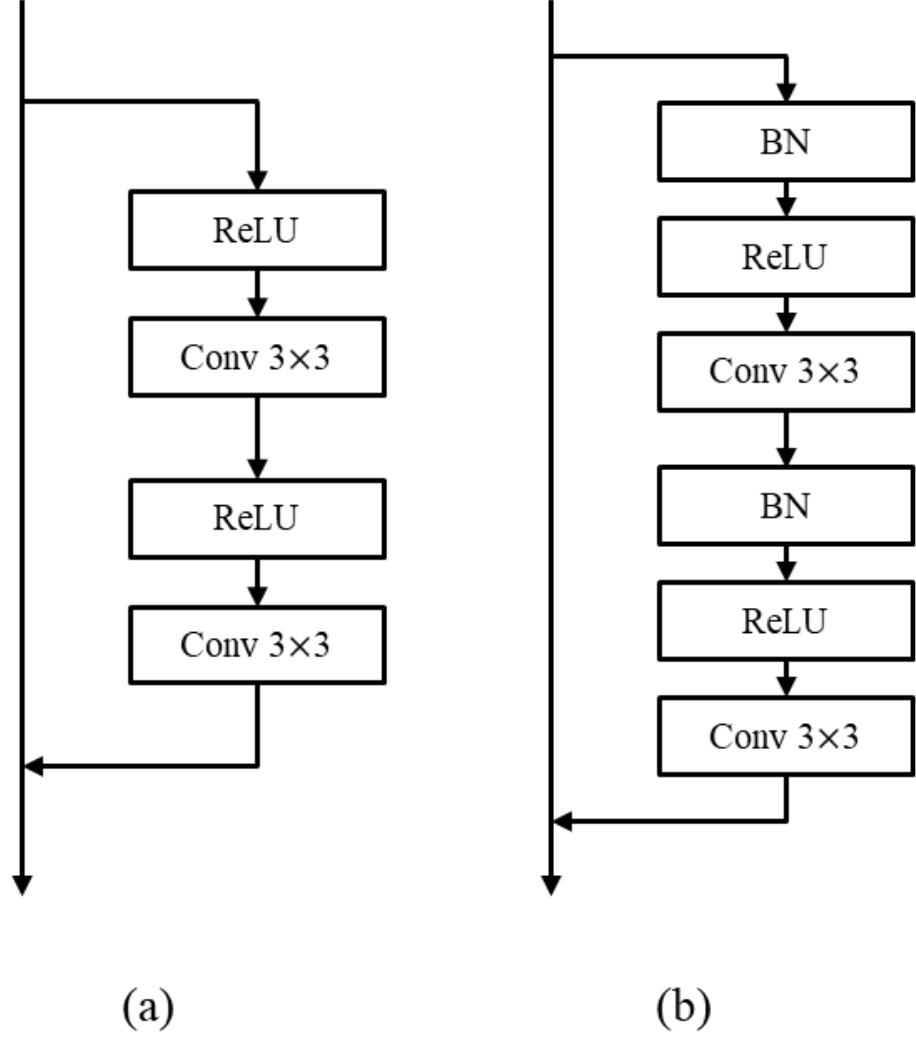}
\caption{Detailed architectures of the ResBlock used in our experiments. (a) ResBlock of the discriminator, (b) ResBlock of the generator.}
\label{fig:fig6}
\end{figure}

\begin{table*}[t]
\caption{Comparison of the proposed method with the traditional GAN on the CIFAR-10, CelebA, LSUN datsets in terms of FID. The bold numbers represent the best performance among the results in each dataset.}
\begin{center}
\begin{tabular}{c | c | c | c | c | c | c}
\hline\hline
& & $N=1$ & $N=2$ & $N=4$ & $N=8$ & $N=16$ \\
\hline
\multirow{4}*{CIFAR-10} & trial 1 & 16.19 & 15.46 & 15.64 & 14.42 & 13.50 \\
 & trial 2 & 16.12 & 15.83 & 15.47 & 15.78 & 14.30 \\
 & trial 3 & 15.75 & 16.23 & 15.78 & 15.58 & 13.90 \\
\cline{2-7}
 & \textbf{Average} & $16.02 \pm 0.23 $ & $15.84 \pm 0.39 $ & $15.63 \pm 0.16 $ & $15.26 \pm 0.73 $ & \textbf{13.90} $\pm$ \textbf{0.40} \\

\hline

\multirow{4}*{CelebA} & trial 1 & 12.62 & 10.86 & 10.59 & 10.36 & 10.20 \\
 & trial 2 & 12.29 & 10.96 & 10.88 & 10.08 & 10.08 \\
 & trial 3 & 11.86 & 11.10 & 11.18 & 10.03 & 10.45 \\
\cline{2-7}
 & \textbf{Average} & $12.26 \pm 0.38 $ & $10.97 \pm 0.12 $ & $10.88 \pm 0.29 $ & $\textbf{10.15} \pm \textbf{0.18} $ & $10.25 \pm 0.19 $ \\

\hline
\multirow{4}*{LSUN} & trial 1 & 20.14 & 17.94 & 17.55 & 16.57 & 18.33 \\
 & trial 2 & 19.19 & 18.66 & 17.16 & 16.89 & 18.50 \\
 & trial 3 & 18.85 & 18.34 & 17.95 & 16.84 & 17.82 \\
\cline{2-7}
 & \textbf{Average} & $19.39 \pm 0.67 $ & $18.31 \pm 0.36 $ & $17.55 \pm 0.39 $ & $\textbf{16.77} \pm \textbf{0.17} $ & $18.22 \pm 0.36 $ \\

\hline\hline
\end{tabular}
\end{center}
\label{table:table3}
\end{table*}

Since all parameters in the generator and the discriminator including the CR module can be differentiated, we performed an optimization using the Adam optimizer~\cite{kingma2014adam}, which is a stochastic optimization method with adaptive estimation of moments. We set the parameters of Adam optimizer, \ie $\beta _1$ and $\beta _2$, to 0 and 0.9, respectively, and set the learning rate to 0.0002. During training procedure, we updated the discriminator five times per each update of the generator. For the CIFAR-10, CelebA, and LSUN datasets, we used a batch size of 64 and trained the generator for 100k iterations, whereas we trained the network 100k iterations with 16 batch sizes to train the CelebA-HQ dataset. For the tiny-ImageNet, we set a batch size as 32 and trained the generator for 500k iterations. Our experiments were conducted on CPU Intel(R) Xeon(R) CPU E3-1245 v5 and GPU RTX 2080 Ti, and implemented in \textit{TensorFlow}.

\section{Experiments}
\label{sec4}
\subsection{Baseline models}
\label{subsec:4.2}
In this work, to produce the image with small size, we employed the generator and discriminator architectures of the leading cGAN scheme~\cite{miyato2018cgans, miyato2018spectral}, as our baseline models. Meanwhile, to generate the image with high-resolution ($256\times256$ and $512\times512$ resolutions), we design the generator and discriminator architectures following the progressive growing of GAN scheme~\cite{karras2017progressive}. The detailed generator and discriminator architectures are presented in Tables~\ref{table:table1} and ~\ref{table:table2}. We employed the multiple residual block~\cite{he2016deep} (ResBlocks) as depicted in Fig.~\ref{fig:fig6}. In the discriminator, we employed the spectral normalization~\cite{miyato2018spectral} for all layers including the proposed CR module. For the discriminator, the down-sampling (average-pooling) is performed after the second convolutional layer, whereas the generator up-sampled the feature maps using a nearest neighbor interpolation prior to the first convolution layer. 

\begin{table}[t]
\caption{Comparison of the proposed method with the traditional GAN on the CelebA and LSUN datasets in terms of FID.}
\begin{center}
\begin{tabular}{c | c | c | c}
\hline\hline
& & Traditional GAN & Proposed method \\
\hline
\multirow{4}*{CelebA} & trial 1 & 20.61 & \textbf{14.22} \\
 & trial 2 & 17.19 & \textbf{14.10} \\
 & trial 3 & 18.07 & \textbf{14.88} \\
\cline{2-4}
 & \textbf{Average} & $18.62 \pm 1.77 $ & $\textbf{14.40} \pm \textbf{0.42}$ \\

\hline
\multirow{4}*{LSUN} & trial 1 & 38.18 & \textbf{34.01} \\
 & trial 2 & 42.19 & \textbf{32.53} \\
 & trial 3 & 43.73 & \textbf{32.89} \\
\cline{2-4}
 & \textbf{Average} & $41.37 \pm 2.87 $ & $\textbf{33.14} \pm \textbf{0.77}$ \\
 
\hline\hline
\end{tabular}
\end{center}
\label{table:table4}
\end{table}

\subsection{Evaluation metric}
\label{subsec:4.3}
In order to evaluate the performance of the generator, in this brief, we employed the principled and comprehensive metric, called frechet inception distance (FID)~\cite{19heusel2017gans}, which measures the visual appearance and diversity of the generated images. The FID can be obtained by calculating the Wasserstein-2 distance between the distribution of the real images, $P_R$, and that of the generated ones, $P_G$, in the feature space obtained via the Inception model~\cite{26szegedy2016rethinking}, which is defined as follows:

\begin{equation}
\label{eq11:FID}
    \textrm{FID}(p,q) = \| \mu_p - \mu_q \|_2^2 + \mathrm{trace}(C_p +C_q - 2(C_p C_q)^{1/2}),
\end{equation}\\
where $ \{\mu_p,C_p \},\{\mu_q,C_q \}$ are the mean and covariance of the samples with distribution $P_R$ and $P_G$, respectively. Lower FID scores indicate better quality of the generated images. In our experiments, we randomly generated 50,000 images for CIFAR-10, LSUN, CelebA, and tiny-ImageNet datasets and 30,000 images for Celeb-HQ dataset. Indeed, there is an alternative approximate measure of image quality, called inceptions score (IS). However, since the IS has some flaws as mentioned in~\cite{barratt2018note, chen2019self}, we employed the FID as the main metric in this brief.\\ 


\begin{table}[t]
\caption{Comparison of the proposed method with the traditional GAN on the tiny-imageNet dataset in terms of FID.}
\begin{center}
\begin{tabular}{c | c | c}
\hline\hline
& Traditional GAN & Proposed method \\
\hline
Unconditional & 66.39 (65.68) & \textbf{63.04 (61.62)} \\
Conditional & 57.17 (52.99) & \textbf{51.84 (50.81)} \\

\hline\hline
\end{tabular}
\end{center}
\label{table:table5}
\end{table}

\subsection{Quantitative comparison}
\label{subsec:4.4}

To demonstrate the advantage of the CR module, we conducted extensive experiments by adjusting the \textit{N} value. Note that $N=1$ is equivalent to the traditional GAN. We trained the network three times to reveal that performance improvement is not caused by lucky weight initialization. Table~\ref{table:table3} shows the average and standard deviation of FID values according to \textit{N} values. As shown in Table~\ref{table:table3}, the CR module significantly improves the performance of GAN. In particular, the performance is improved by increasing the \textit{N} value but FID scores are often unstable when the \textit{N} value is too large. For instance, when $N=16$, the proposed method exhibits better performance compared to the CR module with $N=8$ on the CIFAR-10 dataset but shows slightly lower performance on CelebA and LSUN datasets. However, the proposed method still presents better performance than the traditional GAN. Thus, we confirmed that the proposed method leads the generator successfully to achieve the low FID scores. 

In order to reveal the generalization ability of the proposed method, we conducted additional experiments using other network architectures; we employed DCGAN~\cite{radford2015unsupervised} architectures which consist of multiple transposed convolutional layers in the generator and standard convolutional layers in the discriminator. In our experiments, we applied the spectral normalization to all layers in the discriminator. Based on the previous experimental results in Table~\ref{table:table3}, we set $N$ as 8. As shown in Table~\ref{table:table4}, the proposed method improves the FID score on CelebA and LSUN datasets successfully; these results indicate that the CR module can improve GAN performance regardless of the network architecture. In addition, the performance is improved consistently even training three times in a scratch. These results demonstrate that performance improvement is caused by the CR module, not lucky weight initialization.

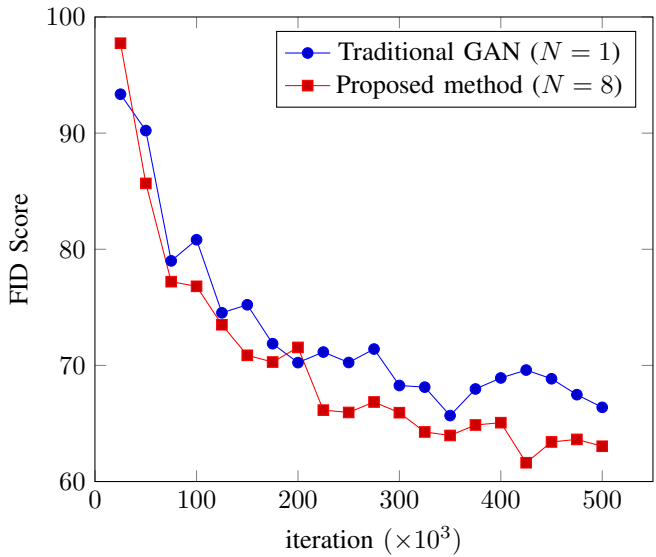
\begin{figure}
\centering
    \begin{tikzpicture}
        \begin{axis}[ ymax = 100, ymin = 60, xmin = -0.2,
                xlabel=iteration $(\times 10^3)$,
                ylabel=FID Score,
                legend pos=north east]
            \addplot+[error bars/.cd,
                      y dir=both, y explicit]
                    coordinates {
                    (25, 93.35) 
                    (50, 90.22) 
                    (75, 79.00) 
                    (100, 80.82)
                    (125, 74.53)
                    (150, 75.22)
                    (175, 71.87)
                    (200, 70.24)
                    (225, 71.15)
                    (250, 70.25)
                    (275, 71.41)
                    (300, 68.27)
                    (325, 68.13)
                    (350, 65.68)
                    (375, 67.97)
                    (400, 68.92)
                    (425, 69.60)
                    (450, 68.85)
                    (475, 67.48)
                    (500, 66.39)
                    };
                \addlegendentry{Traditional GAN ($N=1$)}

            \addplot+[error bars/.cd,
                      y dir=both, y explicit]
                    coordinates {
                    (25, 97.74) 
                    (50, 85.67) 
                    (75, 77.21) 
                    (100, 76.81)
                    (125, 73.50)
                    (150, 70.87)
                    (175, 70.29)
                    (200, 71.55)
                    (225, 66.15)
                    (250, 65.95)
                    (275, 66.85)
                    (300, 65.93)
                    (325, 64.27)
                    (350, 63.97)
                    (375, 64.87)
                    (400, 65.07)
                    (425, 61.62)
                    (450, 63.41)
                    (475, 63.63)
                    (500, 63.04)
                    };
                \addlegendentry{Proposed method ($N=8$)}

        \end{axis}
    \end{tikzpicture}
    \caption{The FID score over the training iteration. The blue and red lines indicate the traditional GAN and the proposed method, respectively.}
\label{fig:fig7}
\end{figure}

To make quantitative results more reliable, we also compared the traditional GAN and the proposed method on tiny-ImageNet with $128\times128$ resolution. Since we already demonstrate that the performance improvement is not caused by lucky weight initialization in Tables~\ref{table:table3} and~\ref{table:table4}, the network is trained a single time in a scratch. As illustrated in Fig.~\ref{fig:fig7} which represents the FID scores over training iterations, the FID score of traditional GAN is saturated earlier than that of the proposed method. As a result, the proposed method exhibits superior performance compared with the traditional GAN. Table~\ref{table:table5} shows the final FID scores of the traditional GAN and the proposed method. The FID score in the bracket indicates a minimum FID score during the training procedure. As described in Table~\ref{table:table5}, even when comparing the minimum FID score, the proposed method shows better performance than traditional GAN method. 



\begin{figure}
\centering
    \begin{tikzpicture}
        \begin{axis}[ ymax = 91, ymin = 45, xmin = -0.2,
                xlabel=iteration $(\times 10^3)$,
                ylabel=FID Score,
                legend pos=north east]
            \addplot+[error bars/.cd,
                      y dir=both, y explicit]
                    coordinates {
                    (25, 90.01) 
                    (50, 82.47) 
                    (75, 71.43) 
                    (100, 65.88)
                    (125, 65.83)
                    (150, 64.77)
                    (175, 61.34)
                    (200, 63.45)
                    (225, 61.68)
                    (250, 58.71)
                    (275, 61.80)
                    (300, 57.11)
                    (325, 59.83)
                    (350, 54.57)
                    (375, 54.41)
                    (400, 57.72)
                    (425, 52.99)
                    (450, 54.43)
                    (475, 54.09)
                    (500, 57.17)

                    };
                \addlegendentry{cGAN~\cite{miyato2018cgans}}

            \addplot+[error bars/.cd,
                      y dir=both, y explicit]
                    coordinates {
                    (25, 88.99) 
                    (50, 72.18) 
                    (75, 73.69) 
                    (100, 64.10)
                    (125, 63.59)
                    (150, 61.43)
                    (175, 59.16)
                    (200, 60.21)
                    (225, 58.92)
                    (250, 58.11)
                    (275, 56.09)
                    (300, 56.13)
                    (325, 55.54)
                    (350, 55.88)
                    (375, 53.58)
                    (400, 53.67)
                    (425, 52.67)
                    (450, 52.78)
                    (475, 50.81)
                    (500, 51.84)
                    };
                \addlegendentry{Proposed method ($N=8$)}

        \end{axis}
    \end{tikzpicture}
    \caption{The FID score over the training iteration. The blue and red lines indicate the traditional cGAN and the proposed method, respectively.}
\label{fig:fig8}
\end{figure}
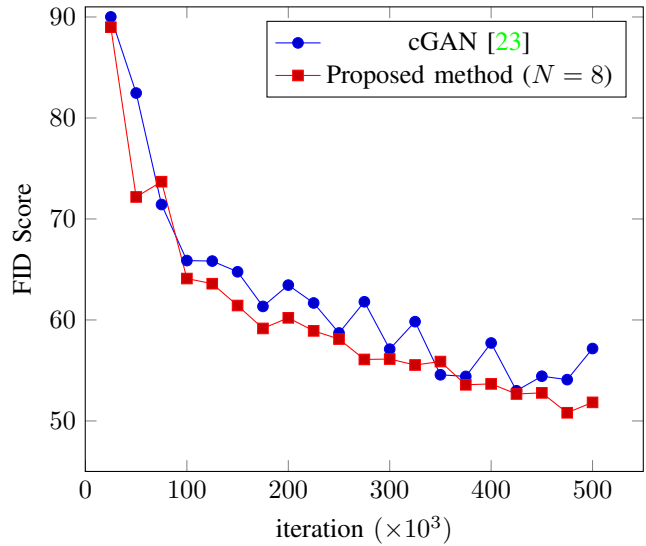

In addition, we conducted additional experiments to validate the effectiveness of the cCR module using tiny-ImageNet consisting of 200 classes. We employed the same baseline model with the experiments of GAN, but replaced the BN in the generator with the conditional batch normalization layer~\cite{dumoulin2017learned}; this conditional-baseline model is equivalent to the model in~\cite{miyato2018cgans} which leads the cGAN scheme. In the proposed method, the CR module in the discriminator is replaced with cCR module. Similar to the trend of experimental results of GAN, as depicted in Fig.~\ref{fig:fig8}, the proposed method achieves better performance than the conventional cGAN framework~\cite{miyato2018cgans}. In addition, as shown in Table~\ref{table:table5}, the final FID score of the proposed method shows better performance compared with the conventional method. These results reveal that the proposed cCR module improves the performance of cGAN by leading the discriminator to consider the dynamic non-overlapped features simultaneously.

\begin{figure*}[t]
\centering
\includegraphics[width=0.85\linewidth]{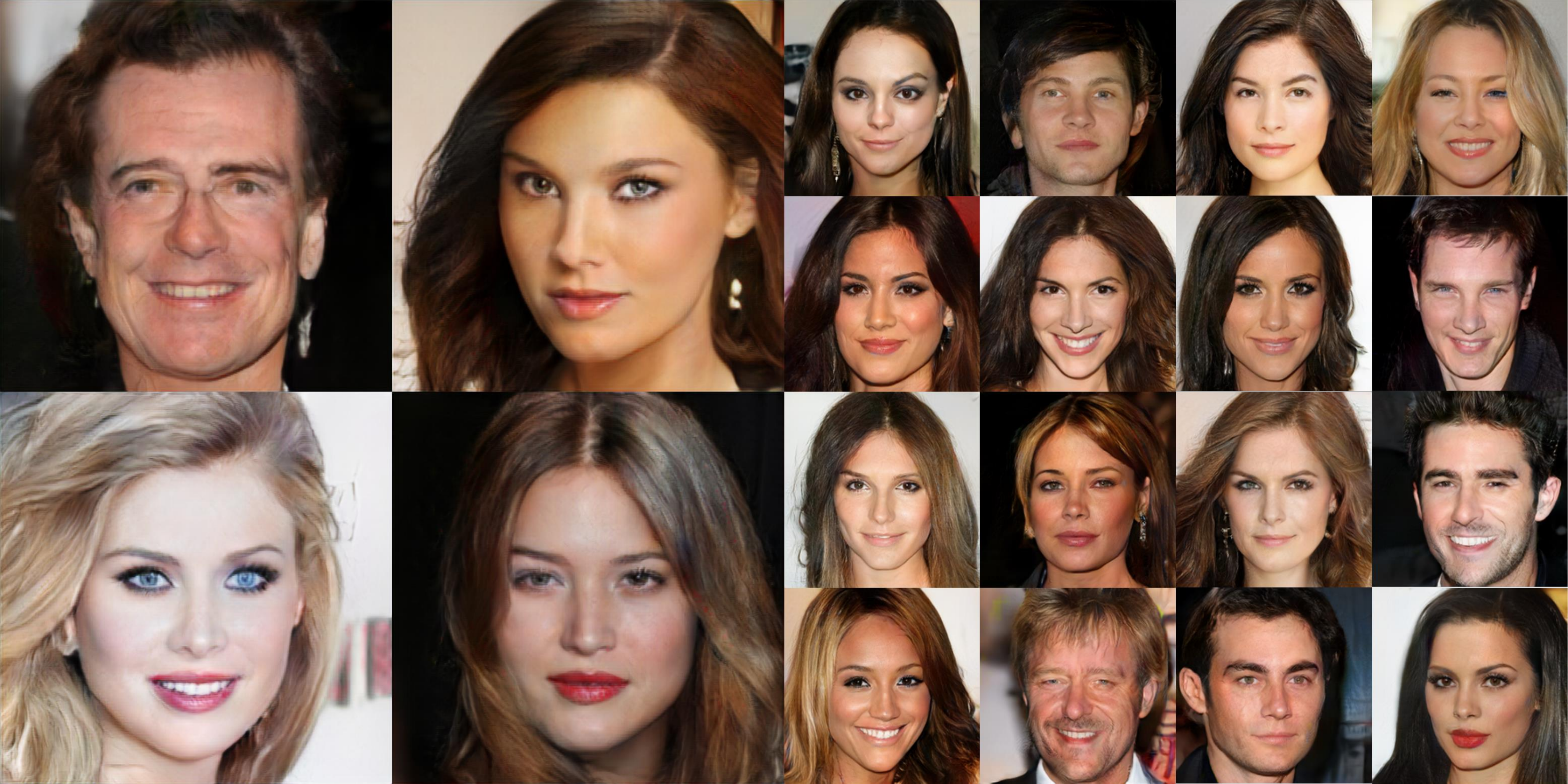}
\caption{Examples of the generated images with $512\times512$ and $256\times256$ resolutions on the Celeb-HQ dataset.}
\label{fig:fig10}
\end{figure*}

It is worth noting that the proposed method additionally requires the $(N-1)\times C_L$ network parameters, where $C_L$ indicates the dimension of the last layer in the discriminator, which are very small numbers compared to the overall discriminator parameters. Thus, the proposed CR module can be added to the discriminator with a marginal training overhead. In addition, since the proposed method is appended after the last convolutional layer of the discriminator, it does not require to modify the discriminator architecture. Indeed, this brief does not intend to design an optimal generator and discriminator architectures for the CR module; there could be another structure that leads to better performance and generates more high-quality images. In contrast, we care more about whether it is possible to improve the performance of GAN by simply adding the CR module to the discriminator.


\subsection{High-resolution image generation}
\label{subsec:4.5}
Furthermore, to demonstrate the generalization ability of the proposed method, we conducted additional experiments that generate a high-resolution image. In our experiments, we trained the networks to generate $256\times256$ and $512\times512$ images using CelebA-HQ dataset. In contrast to the previous work~\cite{karras2017progressive}, we did not employ the progressive growing training scheme. As shown in Table~\ref{table:table6}, the proposed method achieves better performance than the traditional GAN. In addition, Fig.9 shows the examples of generated images. As depicted in Fig.9, the proposed method allows the generator to produce visually pleasing images. These observations demonstrated that the proposed method is effective to generate high-resolution image with high quality. 

\begin{table}[t]
\caption{Comparison of the proposed method with the traditional GAN on the CelebA-HQ dataset in terms of FID.}
\begin{center}
\begin{tabular}{c | c | c}
\hline\hline
& Traditional GAN & Proposed method \\
\hline
$256\times256$ & 24.46 & \textbf{20.93} \\
$512\times512$ & 32.63 & \textbf{30.82} \\
\hline\hline
\end{tabular}
\end{center}
\label{table:table6}
\end{table}

\section{Conclusion}
\label{sec5}
In this brief, we have introduced a straightforward method for improving the performance of GAN. By using the non-overlapped features obtained via the proposed CR module, the discriminator effectively guides the generator during the training procedure, which results in enhancing the ability of the generator. One of the main advantages of the CR module is that it can be readily integrated with the existing discriminator architectures. Moreover, our experiments reveal that, without imposing the training overhead, the discriminator with the CR module significantly improves the performance of the baseline models. In addition, the generalization ability of the proposed method is demonstrated by applying the CR module to high-resolution image. It is expected that the proposed method will be applicable to various applications based on GAN.


\bibliographystyle{IEEEtran}

\bibliography{egbib.bib}

\end{document}